\documentclass[12pt,letterpaper]{article}
\usepackage[a4paper, total={7in, 10in}]{geometry}

\usepackage{graphicx}
\usepackage{helvet}
\usepackage{authblk}
\usepackage{hyperref}
\usepackage{amsmath} 
\usepackage{amssymb} 
\usepackage{orcidlink} 
\usepackage{csvsimple}
\usepackage[super,comma,sort&compress]  
   {natbib}
\usepackage[right]{lineno} \linenumbers
\usepackage[title]{appendix}%

\usepackage[utf8]{inputenc} 
\usepackage[T1]{fontenc}    
\usepackage{url}            
\usepackage{booktabs}       
\usepackage{amsfonts}       
\usepackage{nicefrac}       
\usepackage{microtype}      
\usepackage{xcolor}         
\usepackage{multirow}
\usepackage{threeparttable}
\usepackage{longtable}
\usepackage{amsmath}
\usepackage{subcaption}
\usepackage{hyperref}

\usepackage{xcolor}
\hypersetup{
    colorlinks=true,
    linkcolor=blue,
    filecolor=magenta,      
    urlcolor=blue, 
    pdftitle={Your Title},
}
\usepackage{makecell}
\usepackage{placeins}
\usepackage[utf8]{inputenc} 
\usepackage{booktabs}      
\usepackage{siunitx}       
\usepackage{caption}       
\usepackage{tabularx}
\usepackage{listings}    
\usepackage{pifont}
\usepackage{array}
\usepackage{makecell}
\theadset{\bfseries}

\makeatletter
\renewcommand{\maketitle}{\bgroup\setlength{\parindent}{0pt}
\begin{flushleft}
  \textbf{\@title}
  
  \@author
\end{flushleft}\egroup}
\makeatother

\title{AnyECG-Lab: An Exploration Study of Fine-tuning an ECG Foundation Model to Estimate Laboratory Values from Single-Lead ECG Signals}
\date{}

\author[1,2]{Yujie Xiao}
\author[1,2]{Gongzhen Tang}
\author[1,2]{Wenhui Liu}
\author[1,2]{Jun Li}
\author[1,2,3]{Guangkun Nie}
\author[1,2]{Zhuoran Kan}
\author[6]{Deyun Zhang}
\author[7]{Qinghao Zhao}
\author[1,2,4,5,*,\orcidlink{0000-0001-7521-5127}]{Shenda Hong}

\affil[1]{Institute of Medical Technology, Peking University Health Science Center, Beijing, China}
\affil[2]{National Institute of Health Data Science, Peking University, Beijing, China}
\affil[3]{School of Intelligence Science and Technology, Peking University, Beijing, China}
\affil[4]{Institute for Artificial Intelligence, Peking University, Beijing, China}
\affil[5]{State Key Laboratory of Vascular Homeostasis and Remodeling, NHC Key Laboratory of Cardiovascular Molecular Biology and Regulatory Peptides, Peking University, Beijing, China}
\affil[6]{HeartVoice Medical Technology, Hefei, China}
\affil[7]{Department of Cardiology, Peking University People’s Hospital, Beijing, China}

\affil[*]{Correspondence: hongshenda@pku.edu.cn}

\begin{document}

\maketitle

\section*{ABSTRACT}
Timely access to laboratory values is critical for clinical decision-making, yet current approaches rely on invasive venous sampling and are intrinsically delayed. Electrocardiography (ECG), as a non-invasive and widely available signal, offers a promising modality for rapid laboratory estimation. Recent progress in deep learning has enabled the extraction of latent hematological signatures from ECGs. However, existing models are constrained by low signal-to-noise ratios, substantial inter-individual variability, limited data diversity, and suboptimal generalization, especially when adapted to low-lead wearable devices. In this work, we conduct an exploratory study leveraging transfer learning to fine-tune ECGFounder, a large-scale pre-trained ECG foundation model, on the Multimodal Clinical Monitoring in the Emergency Department (MC-MED) dataset from Stanford. We generated a corpus of more than 20 million standardized ten-second ECG segments to enhance sensitivity to subtle biochemical correlates. On internal validation, the model demonstrated strong predictive performance (area under the curve above 0.65) for thirty-three laboratory indicators, moderate performance (between 0.55 and 0.65) for fifty-nine indicators, and limited performance (below 0.55) for sixteen indicators. This study provides an efficient artificial-intelligence driven solution and establishes the feasibility scope for real-time, non-invasive estimation of laboratory values.

\section*{KEYWORDS}
Electrocardiogram, Laboratory Values, Artificial Intelligence, Deep Learning

\section*{INTRODUCTION}
In clinical practice, the diagnosis and treatment decisions of many diseases are highly dependent on the results of laboratory test values\cite{aghaei2024need,gao2024laboratory}. Therefore, timely and accurate acquisition of laboratory values is of great clinical significance. However, the current commonly used method of collecting laboratory values mainly relies on the collection of venous blood samples\cite{lima2015laboratory} and sending them to the laboratory for analysis. This process is not only invasive and may cause adverse events such as anemia in patients\cite{whitehead2019interventions}, but there is often a significant time delay in the entire process from blood collection and specimen transportation to laboratory analysis\cite{timedelay}. The above factors make traditional laboratory testing methods have obvious limitations in certain clinical scenarios that require rapid decision-making or hope to achieve large-scale, real-time monitoring.

ECG, as an inexpensive, noninvasive, and rapidly acquired method for detecting physiological signals, has been widely used in clinical practice\cite{ecguseinclinical}. ECG can reflect changes in the electrical activity of the cardiovascular system, indirectly reflecting the physiological state of the body, and offers new possibilities to acquire non-invasiveve information related to laboratory values. In recent years, with the rapid development of deep learning in the fields of medical imaging and physiological signal analysis\cite{chen2024towards,li2025experteeg}, artificial intelligence-based ECG analysis (AI-ECG) has been widely used for disease prediction and risk assessment\cite{aiecg1,aiecg2,khurshid2022ecgai3,strodthoff2020deepaiecg4,jiang2024deepai5}. Previous studies have demonstrated that AI-ECG models can predict diseases associated with abnormal laboratory values\cite{lin2022point}. For example, Chiu et al. used an AI-ECG model to predict hyperkalemia from retrospective data, achieving good performance\cite{chiu2024serum}. This suggests that deep learning models can automatically identify potential features associated with changes in laboratory values from ECG signals.

However, the current use of AI-ECG for estimating laboratory values still has certain limitations. Research using AI-ECG methods to predict laboratory values is still in its early stages. Although some studies such as Cardiolab\cite{alcaraz2024cardiolab} show promising results in estimating certain values like NTporBNP, key challenges remain. Specifically, the generalization capabilities of the models and the influence of individual differences require systematic validation. Furthermore, existing research is often based on limited datasets, and most of the studies use 12-lead ECG as input\cite{avula2023clinical}, making direct transfer to low-lead scenarios like wearable devices difficult\cite{presacan2025evaluating}. Therefore, the feasibility of comprehensively predicting multiple laboratory parameters from ECGs requires further in-depth research and verification.

Based on this, this study intends to conduct exploratory work on AI-ECG to predict laboratory values from retrospective data: the study aims to screen out examination items that are significantly correlated with ECG signals from all common and clinically significant laboratory values through training on a very large retrospective data set. Based on the publicly available ECG foundation model, ECGFounder\cite{li2025electrocardiogram,hong2020holmes}, the study systematically built an AI-ECG model for predicting laboratory values using large, real emergency cohort data and reasonably evaluated the performance of the model. The study aims to build a large-scale screening framework to provide new insights and ideas for the subsequent use of ECG to predict laboratory values.

\section*{RESULTS}
\subsection*{Baseline Characteristics of the Data}
This study collected baseline characteristics for the full dataset, training set, and test set. These characteristics included demographic data (age, sex, race, ethnicity) and vital signs (temperature, heart rate, respiratory rate, blood oxygen saturation, systolic and diastolic blood pressure). The following Table~\ref{tab:baseline} summarizes the baseline characteristics of the data.

\begin{table}[htbp]
\renewcommand{\arraystretch}{1.3}
\caption{Baseline Characteristics of the Data} \label{tab:baseline} 
\resizebox{\textwidth}{!}{
\begin{tabular}{l|c|c|c}
\hline
\textbf{Demographics} & \textbf{All} & \textbf{Train} & \textbf{Test} \\
\hline
\textbf{Age (years)} &  &  &  \\
Mean ± Standard Deviation & 59.3 ± 19.8 & 59.3 ± 19.8 & 59.6 ± 19.9 \\
Median & 61.0 [44.0, 76.0] & 61.0 [44.0, 76.0] & 62.0 [44.0, 76.0] \\
\hline
\textbf{Sex} & & & \\
Male & 15327 & 12890 & 3921 \\
Female & 16478 & 13797 & 4219 \\
\hline
\textbf{Race} & & & \\
White & 13966 & 11689 & 3572 \\
Black or African American & 1683 & 1434 & 478 \\
Asian & 5791 & 4860 & 1457 \\
Native Hawaiian or Other Pacific Islander & 599 & 507 & 170 \\
American Indian or Alaska Native & 76 & 59 & 25 \\
Other & 9474 & 7973 & 2377 \\
Declines to State & 158 & 123 & 44 \\
Unknown & 58 & 42 & 17 \\
\hline
\textbf{Ethnicity} & & & \\
Non-Hispanic/Non-Latino & 24116 & 20242 & 6165 \\
Hispanic/Latino & 7327 & 6158 & 1887 \\
Unknown & 157 & 124 & 36 \\
Declines to State & 205 & 163 & 52 \\
\hline
\textbf{Vital Signs} & & & \\
\textbf{Temperature (T) (°C)} & & & \\
Mean ± Standard Deviation & 36.7 ± 0.54 & 36.7 ± 0.55 & 36.7 ± 0.55 \\
Median & 36.7 [36.5, 36.9] & 36.7 [36.5, 36.9] & 36.7 [36.5, 36.9] \\
\hline
\textbf{Heart Rate (HR) (bpm)} & & & \\
Mean ± Standard Deviation & 88.6 ± 20.7 & 88.6 ± 20.7 & 88.6 ± 20.7 \\
Median & 86.0 [74.0, 101.0] & 86.0 [74.0, 101.0] & 86.0 [74.0, 101.0] \\
\hline
\textbf{Respiratory Rate (RR) (breaths/min)} & & & \\
Mean ± Standard Deviation & 18.6 ± 3.63 & 18.6 ± 3.60 & 18.6 ± 3.6 \\
Median & 18.0 [16.0, 20.0] & 18.0 [16.0, 20.0] & 18.0 [16.0, 20.0] \\
\hline
\textbf{Oxygen Saturation (SpO2) (\%)} & & & \\
Mean ± Standard Deviation & 97.9 ± 2.82 & 97.9 ± 2.82 & 97.9 ± 2.82 \\
Median & 99.0 [97.0, 100.0] & 99.0 [97.0, 100.0] & 99.0 [97.0, 100.0] \\
\hline
\textbf{Systolic Blood Pressure (SBP) (mmHg)} & & & \\
Mean ± Standard Deviation & 136.8 ± 24.6 & 136.8 ± 24.6 & 136.8 ± 24.6 \\
Median & 135.0 [120.0, 151.0] & 136.0 [120.0, 151.0] & 136.0 [120.0, 151.0] \\
\hline
\textbf{Diastolic Blood Pressure (DBP) (mmHg)} & & & \\
Mean ± Standard Deviation & 79.2 ± 16.5 & 79.2 ± 16.5 & 79.2 ± 16.6 \\
Median & 79.0 [68.0, 89.0] & 79.0 [68.0, 89.0] & 79.0 [68.0, 89.0] \\
\hline
\end{tabular}
}
\end{table}

\subsection*{Performance of AnyECG-Lab}
An exploratory analysis was conducted to assess the feasibility of using ECG signals to predict multiple laboratory values. The overall prediction results of the model are shown in Table~\ref{tab:long_auc_results}.

\begin{scriptsize}
\renewcommand{\arraystretch}{1.3}
\begin{longtable}{l|c|c|c|c} 

\caption{Model prediction AUC of various laboratory values (1 hour window)} \label{tab:long_auc_results} \\

\hline
\textbf{Laboratory Name} & \textbf{Range} & \textbf{All data num} & \textbf{Positivte data num} & \textbf{AUC with CI (1 hours)} \\
\hline
\endfirsthead

\multicolumn{3}{c}{{\small\bfseries \tablename\ \thetable{}  (Continued from previous page)}} \\
\hline
\textbf{Laboratory Name} & \textbf{Range} & \textbf{All data num} & \textbf{Positivte data num} & \textbf{AUC with CI (1 hours)} \\
\hline
\endhead 

\hline

\multicolumn{3}{c}{{\small\itshape Continue to next page}} \\
\endfoot
\hline
\endlastfoot
LDH, TOTAL & <140.0 (U/L) & 124490 & 1486 & 0.851 [0.827-0.873] \\
BNP, NT-PRO & >300.0 (pg/mL) & 1340317 & 855688 & 0.783 [0.781-0.785] \\
CALCIUM, IONIZED & >1.3 (mmol/L) & 126551 & 16645 & 0.755 [0.745-0.764] \\
BLOOD UREA NITROGEN (BUN) & <7.0 (mg/dL) & 6921636 & 264614 & 0.714 [0.712-0.716] \\
POTASSIUM & <3.5 (mmol/L) & 6773875 & 546014 & 0.713 [0.712-0.715] \\
MONOCYTE \% (AUTO DIFF) & <2.0 (\%) & 6665742 & 66890 & 0.713 [0.708-0.717] \\
CREATININE & >1.3 (mg/dL) & 6736140 & 1195026 & 0.709 [0.708-0.710] \\
FERRITIN & <13.0 (ng/mL) & 109535 & 8098 & 0.709 [0.698-0.721] \\
BLOOD UREA NITROGEN (BUN) & >20.0 (mg/dL) & 6921636 & 2065797 & 0.706 [0.705-0.707] \\
OSMOLALITY, SERUM & <275.0 (mOsm/kg) & 405950 & 72436 & 0.695 [0.690-0.700] \\
RED CELL DISTRIBUTION WIDTH (RDW) & <11.5 (\%) & 7425805 & 53719 & 0.689 [0.685-0.693] \\
ALBUMIN & >5.0 (g/dL) & 6565749 & 109383 & 0.688 [0.685-0.692] \\
TRANSFERRIN SATURATION & >50.0 (\%) & 57384 & 4622 & 0.686 [0.661-0.708] \\
ALBUMIN & <3.5 (g/dL) & 6565749 & 988415 & 0.683 [0.682-0.684] \\
IRON>TIBC RATIO & >0.5 (RATIO) & 57584 & 4822 & 0.681 [0.658-0.703] \\
CORRECTED CALCIUM & >10.5 (mg/dL) & 6545135 & 113612 & 0.680 [0.677-0.683] \\
POTASSIUM & >5.0 (mmol/L) & 6773875 & 555336 & 0.678 [0.677-0.680] \\
SODIUM & >145.0 (mmol/L) & 6930467 & 70155 & 0.672 [0.667-0.676] \\
RED CELL DISTRIBUTION WIDTH (RDW) & >14.5 (\%) & 7425805 & 2438727 & 0.668 [0.667-0.669] \\
PLATELET COUNT (PLT) & >450.0 (K/uL) & 7432728 & 303285 & 0.668 [0.665-0.670] \\
PROTHROMBIN TIME & >13.5 (seconds) & 2675448 & 1869380 & 0.665 [0.664-0.667] \\
RED BLOOD CELLS (RBC) & <3.5 (MIL/uL) & 7432560 & 1312467 & 0.664 [0.663-0.665] \\
AST (SGOT) & <10.0 (U/L) & 6553480 & 16061 & 0.664 [0.655-0.673] \\
FREE THYROXINE & <0.8 (ng/dL) & 130429 & 7463 & 0.664 [0.648-0.678] \\
ALKALINE PHOSPHATASE & >150.0 (U/L) & 6555594 & 758397 & 0.661 [0.659-0.662] \\
D-DIMER, ELISA & >500.0 (ng/mL FEU) & 382224 & 190573 & 0.661 [0.657-0.665] \\
NEUTROPHIL \% (AUTO DIFF) & >75.0 (\%) & 6665742 & 2138742 & 0.660 [0.659-0.661] \\
ANION GAP & >12.0 (mmol/L) & 6733561 & 2448621 & 0.659 [0.658-0.660] \\
LYMPHOCYTE \% (AUTO DIFF) & <20.0 (\%) & 6665742 & 3395401 & 0.656 [0.655-0.657] \\
CHLORIDE & <98.0 (mmol/L) & 6736101 & 1351359 & 0.654 [0.653-0.655] \\
C-REACTIVE PROTEIN (CRP) & >1.0 (mg/dL) & 325961 & 222728 & 0.652 [0.648-0.656] \\
WHITE BLOOD CELLS (WBC) & >10.0 (K/uL) & 7440146 & 2050853 & 0.651 [0.650-0.652] \\
D-DIMER & >0.5 (ug/mL FEU) & 280234 & 179415 & 0.650 [0.645-0.654] \\
CHOLESTEROL, HDL & <40.0 (mg/dL) & 155336 & 38693 & 0.649 [0.641-0.657] \\
CALCIUM & >10.5 (mg/dL) & 6750020 & 115388 & 0.648 [0.645-0.651] \\
ESR (AUTOMATED) & >29.0 (mm/hr) & 223325 & 110745 & 0.648 [0.644-0.653] \\
CHOLESTEROL/HDL & >5.0 (ratio) & 155136 & 21487 & 0.648 [0.640-0.656] \\
CREATININE & <0.6 (mg/dL) & 6736140 & 959065 & 0.647 [0.646-0.649] \\
HEMATOCRIT (HCT) & >50.0 (\%) & 7437170 & 122931 & 0.644 [0.641-0.647] \\
RED BLOOD CELLS (RBC) & >5.7 (MIL/uL) & 7432560 & 160306 & 0.644 [0.641-0.647] \\
AMMONIA, PLASMA & >50.0 (umol/L) & 160167 & 52046 & 0.644 [0.638-0.650] \\
HEMATOCRIT (HCT) & <36.0 (\%) & 7437170 & 2342373 & 0.641 [0.640-0.642] \\
CALCIUM & <8.5 (mg/dL) & 6750020 & 719235 & 0.641 [0.639-0.642] \\
MEAN CORPUSCULAR VOLUME (MCV) & >100.0 (fL) & 7430763 & 578388 & 0.641 [0.639-0.642] \\
GLOBULIN & >3.5 (g/dL) & 6545109 & 1168742 & 0.640 [0.639-0.642] \\
PHOSPHORUS & >4.5 (mg/dL) & 333008 & 39754 & 0.640 [0.634-0.646] \\
PROTEIN, TOTAL & <6.0 (g/dL) & 6555383 & 401859 & 0.638 [0.636-0.640] \\
BILIRUBIN, TOTAL & >1.2 (mg/dL) & 6346974 & 528806 & 0.637 [0.636-0.639] \\
OSMOLALITY, SERUM & >295.0 (mOsm/kg) & 405950 & 179235 & 0.637 [0.633-0.641] \\
PART. THROMBOPLASTIN TIME & >35.0 (seconds) & 2551279 & 664536 & 0.636 [0.634-0.637] \\
GLUCOSE & >140.0 (mg/dL) & 6921135 & 1659736 & 0.635 [0.634-0.636] \\
HIGH SENSITIVITY TROPONIN & >14.0 (ng/L) & 1133960 & 465349 & 0.632 [0.629-0.634] \\
SODIUM & <135.0 (mmol/L) & 6930467 & 1470355 & 0.630 [0.629-0.631] \\
PLATELET COUNT (PLT) & <150.0 (K/uL) & 7432728 & 927521 & 0.627 [0.626-0.629] \\
CO2 & <22.0 (mmol/L) & 6735205 & 1278366 & 0.625 [0.623-0.626] \\
CHOLESTEROL, TOTAL & >200.0 (mg/dL) & 155926 & 44633 & 0.624 [0.617-0.631] \\
IMMATURE GRANULOCYTE \% (AUTODIFF) WAM & >1.0 (\%) & 6656289 & 554831 & 0.623 [0.621-0.625] \\
PROTEIN, TOTAL & >8.0 (g/dL) & 6555383 & 480463 & 0.620 [0.618-0.622] \\
CHOLESTEROL, DIRECT LDL & >130.0 (mg/dL) & 149478 & 28885 & 0.620 [0.611-0.628] \\
CALCIUM, IONIZED & <1.1 (mmol/L) & 126551 & 26265 & 0.616 [0.607-0.624] \\
AST (SGOT) & >40.0 (U/L) & 6553480 & 1727314 & 0.611 [0.610-0.612] \\
ALT (SGPT) & >40.0 (U/L) & 6541309 & 1391991 & 0.611 [0.609-0.612] \\
MAGNESIUM & <1.7 (mg/dL) & 1191156 & 92560 & 0.607 [0.602-0.612] \\
CHOLESTEROL, NON-HDL (CALCULATED) & >130.0 (mg/dL) & 155136 & 53447 & 0.607 [0.600-0.613] \\
EOSINOPHIL \% (AUTO DIFF) & >5.0 (\%) & 6665742 & 419455 & 0.606 [0.605-0.608] \\
LYMPHOCYTE \% (AUTO DIFF) & >40.0 (\%) & 6665742 & 384520 & 0.605 [0.603-0.607] \\
UIBC & >450.0 (ug/dL) & 56854 & 2674 & 0.602 [0.572-0.629] \\
TRIGLYCERIDE & >150.0 (mg/dL) & 163468 & 52725 & 0.601 [0.595-0.608] \\
TIBC & >450.0 (ug/dL) & 59162 & 4391 & 0.601 [0.582-0.622] \\
ANION GAP & <8.0 (mmol/L) & 6733561 & 224101 & 0.600 [0.597-0.603] \\
ETHANOL.SER/PLAS & >10.0 (mg/dL) & 299913 & 115607 & 0.600 [0.595-0.604] \\
LACTATE.WHOLE BLD & >2.0 (mmol/L) & 72645 & 35838 & 0.594 [0.585-0.603] \\
CO2 & >29.0 (mmol/L) & 6735205 & 300767 & 0.590 [0.588-0.593] \\
CORRECTED CALCIUM & <8.5 (mg/dL) & 6545135 & 467948 & 0.587 [0.585-0.589] \\
FREE THYROXINE & >1.8 (ng/dL) & 130429 & 17717 & 0.583 [0.573-0.593] \\
CHOL/LDLD RATIO & <3.0 (ratio) & 149358 & 143473 & 0.581 [0.569-0.593] \\
CHLORIDE & >106.0 (mmol/L) & 6736101 & 542977 & 0.580 [0.578-0.582] \\
ALKALINE PHOSPHATASE & <40.0 (U/L) & 6555594 & 101703 & 0.580 [0.576-0.584] \\
MEAN CORPUSCULAR VOLUME (MCV) & <80.0 (fL) & 7430763 & 404961 & 0.577 [0.575-0.579] \\
GLUCOSE & <70.0 (mg/dL) & 6921135 & 60403 & 0.575 [0.570-0.579] \\
OSMOLAL GAP, SER/PLAS & >10.0 (mOsm/kg) & 298535 & 12440 & 0.575 [0.564-0.586] \\
TSH & >4.2 (uIU/mL) & 708236 & 109519 & 0.569 [0.564-0.573] \\
PROCALCITONIN & >0.05 (ng/mL) & 263100 & 210723 & 0.566 [0.561-0.572] \\
BASOPHIL \% (AUTO DIFF) & >1.0 (\%) & 6665742 & 407345 & 0.564 [0.562-0.566] \\
PHOSPHORUS & <2.5 (mg/dL) & 333008 & 55144 & 0.564 [0.558-0.570] \\
LDH, TOTAL & >280.0 (U/L) & 124490 & 79582 & 0.562 [0.555-0.569] \\
FERRITIN & >400.0 (ng/mL) & 109535 & 47882 & 0.562 [0.554-0.569] \\
MONOCYTE \% (AUTO DIFF) & >10.0 (\%) & 6665742 & 1411662 & 0.561 [0.559-0.562] \\
OSMOLALITY, CALCULATED & >295.0 (mOsm/kg) & 299713 & 160253 & 0.559 [0.554-0.563] \\
LIPASE & >160.0 (U/L) & 1578641 & 49403 & 0.556 [0.550-0.561] \\
WHITE BLOOD CELLS (WBC) & <4.0 (K/uL) & 7440146 & 417908 & 0.552 [0.550-0.554] \\
TROPONIN I & >0.04 (ng/mL) & 813572 & 490682 & 0.549 [0.546-0.551] \\
CK, TOTAL & >250.0 (U/L) & 324911 & 95073 & 0.543 [0.538-0.547] \\
OSMOLALITY, CALCULATED & <275.0 (mOsm/kg) & 299713 & 8003 & 0.543 [0.527-0.557] \\
TIBC & <250.0 (ug/dL) & 59162 & 22940 & 0.530 [0.517-0.542] \\
NEUTROPHIL \% (AUTO DIFF) & <40.0 (\%) & 6665742 & 130901 & 0.528 [0.525-0.532] \\
MAGNESIUM & >2.2 (mg/dL) & 1191156 & 310497 & 0.528 [0.525-0.530] \\
PART. THROMBOPLASTIN TIME & <25.0 (seconds) & 2551279 & 180207 & 0.527 [0.524-0.530] \\
UIBC & <250.0 (ug/dL) & 56854 & 34252 & 0.522 [0.511-0.532] \\
GLOBULIN & <2.0 (g/dL) & 6545109 & 117373 & 0.515 [0.512-0.519] \\
OSMOLAL GAP, SER/PLAS & <0.0 (mOsm/kg) & 298535 & 180954 & 0.495 [0.491-0.500] \\
TSH & <0.27 (uIU/mL) & 708236 & 24555 & 0.495 [0.487-0.503] \\
TRANSFERRIN SATURATION & <20.0 (\%) & 57384 & 35313 & 0.455 [0.443-0.467] \\
CK, TOTAL & <30.0 (U/L) & 324911 & 14597 & 0.440 [0.431-0.450] \\
IRON, TOTAL & <50.0 (ug/dL) & 62950 & 37654 & 0.413 [0.404-0.424] \\
IRON, TOTAL & >175.0 (ug/dL) & 62950 & 2853 & 0.407 [0.392-0.421] \\
PROTHROMBIN TIME & <11.0 (seconds) & 2675448 & 335 & 0.398 [0.355-0.446] \\
IRON>TIBC RATIO & <0.15 (RATIO) & 57584 & 30436 & 0.394 [0.383-0.405] \\
\end{longtable}
\end{scriptsize}

    
    
The results showed that among the 108 indicators evaluated with laboratory examination time and ECG examination time within one hour, the predictive ability of the model showed obvious stratification:
\begin{itemize}
    \item \textbf{Strong Predictive Ability ($AUC \ge 0.65$)}: A total of 33 indicators, including "Low LDH, TOTAL," "High BNP, NT-PRO," and "High CALCIUM, IONIZED," achieved an AUC of 0.65 or higher. This indicates a strong correlation between ECG signals and these indicators, and the model effectively captures relevant features. The model performed best in predicting "Low LDH, TOTAL," with an AUC of 0.851 (95\% CI: 0.827-0.873).
    \item \textbf{Moderate Predictive Ability ($0.55 \le AUC < 0.65$):} A total of 59 indicators had an AUC between 0.55 and 0.65, such as "high D-DIMER" and "low CHOLESTEROL, HDL." This suggests that relevant information may exist in the ECG signal, but it is difficult to extract, and the model's predictive ability is relatively limited.
    \item \textbf{Weak/No Predictive Ability ($AUC < 0.55$):} The AUCs of the remaining 16 indicators (such as "high TROPONIN I") are lower than 0.55, indicating that the ECG signals are weakly correlated with these indicators and the model cannot make accurate predictions.
\end{itemize}


The complete evaluation results for sensitivity, specificity, and F1-score across 108 laboratory values and three time windows (1 h, 30 min, and 15 min) can be found at \url{https://github.com/PKUDigitalHealth/AnyECG-Lab}.

\section*{DISCUSSION}
In clinical practice, laboratory test values are particularly critical for the diagnosis and treatment of diseases. However, the commonly used methods for laboratory examinations currently rely heavily on venous blood sampling, but due to its invasiveness and significant time delay, it is difficult to meet the needs of rapid examinations in clinical scenarios. In recent years, with the widespread application of AI-ECG in clinical practice, it has provided new ideas for the non-invasive and rapid detection of laboratory values. However, the information related to laboratory values contained in ECG signals is weak and has obvious individual differences, which limits the application of ECG in the prediction of a wide range of laboratory values.

To address these issues, this study systematically constructed an AI-ECG model for predicting all common laboratory values based on a large-scale retrospective dataset. The main contribution of our study is to provide new ideas on how to use AI-ECG to predict laboratory values, reveal the strength of the correlation between various values and ECG signals, and provide an empirical basis for future research.



The results of this exploratory study lay the foundation for subsequent research. Future work will gradually expand in depth and breadth:



    
\begin{enumerate}
    \item \textbf{Large-scale multicenter validation:}
    Given the single-center exploratory nature of this study, a top priority in the future is to construct a large-scale, multicenter prospective or retrospective cohort to validate the robustness and generalizability of the current model in different populations, devices, and medical practices.
    
    \item \textbf{Clinical relevance and validation of key laboratory values:}
    We will conduct further clinical validation of the laboratory values with high AUC revealed in this study, focusing on analyzing the immediate correlation between ECG-based dynamic prediction results and established clinical disease severity scores to evaluate their potential clinical application as new digital biomarkers.

    \item \textbf{Research from classification to regression:}
    While current research focuses on classification tasks, future work will shift to regression tasks, using ECG to directly predict the specific laboratory values. This research aims to reveal the quantitative relationship between ECG signal morphology and fluctuations in laboratory values, providing deeper insights into the underlying electrophysiological-biochemical coupling mechanisms and providing a theoretical basis for non-invasive continuous monitoring.
    
\end{enumerate}


\section*{METHODS}

\subsection*{Dataset}

Research utilizd Multimodal Clinical Monitoring in the Emergency Department (MC-MED) dataset collected by Stanford Adult Emergency Department, containing multimodal clinical and physiological data of 118,385 adult emergency department patients who visited the Stanford Adult Emergency Department monitoring bed between September 2020 and September 2022\cite{kansal2025multimodal}. ECG data, as part of the data record, records the patient's second lead ECG with a sampling frequency of 500Hz. Laboratory values record the results of each test item of each laboratory examination for all patients. The dataset is open to the public in a completely anonymous manner, and CSN (Visit identifier) is used to connect waveform data and other patient data, so that researchers can use the dataset to match ECG with laboratory test results.

Study first performed systematic preprocessing at the data level. We first performed a rapid statistical analysis of all laboratory values, using the statistical principle of "the number of unique CSNs that match the ECG waveform" as the effective count for each indicator. Subsequently, to ensure the clinical representativeness and data stability of the analysis indicators, we screened the laboratory values, retaining only common test items that appeared frequently in the overall data set to construct a core indicator set for subsequent modeling and analysis. Under the guidance of professional clinicians, we screened out test items such as "COVID-19" and "barbiturates in urine" that were of low clinical significance, binary (yes or no), and whose sampling source was not blood. Finally, 108 indicators were selected as target laboratory values. In this study, the high and low value intervals of each laboratory indicator were obtained by calling the large language model (LLM), and the resulting intervals were reviewed by professional clinicians.

Then we considered laboratory test results collected at the same time within each CSN as a single blood test event. For each blood test, we extracted target laboratory values and coded them numerically according to the reference normal range: if the parameter was not tested, it was recorded as $-1$; if tested and the result was abnormal, it was recorded as $1$; if tested and within the normal range, it was recorded as $0$. Finally, the codes for the 108 target parameters were organized into a one-hot binary vector. 

In research, a total of 269,029 blood test records were obtained, involving 45,770 independent CSNs. Subsequently, we performed a hierarchical partitioning based on the CSN, dividing the samples into training and test sets in a 4:1 ratio. For the training set, we used the timestamp of each laboratory blood test as the center point, retrieved all ECG recordings within a 1-hour time window before and after it, and segmented the ECG signals into 10-second segments. For the test set, we used the same strategy but constructed three time windows centered on the blood test timestamp: 1 hour, 30 minutes, and 15 minutes. For each blood test, we randomly selected 200 10-second ECG segments from the corresponding time window as the model input for that test. Finally, the training set contains 36,616 CSNs, totaling 15,580,163 10-second ECG segments; the test set contains 9,154 CSNs, of which the 1-hour window, 30-minute window, and 15-minute window contain 3,934,378, 3,026,394, and 2,043,526 10-second ECG segments, respectively.

\subsection*{Model Development}

In model level, we leveraged transfer learning to develop our AI-ECG model. Specifically, we made use of an existing ECG foundation model for downstream fine-tuning to predict whether a target test item was abnormal or not. The base model, ECGFounder, is a ResNet-based architecture pre-trained on a large-scale ECG dataset, capable of capturing rich temporal and spatial features associated with cardiac signals\cite{li2025electrocardiogram,hong2020holmes}. By using transfer learning, the model maintains its original ECG feature extraction capabilities while also enabling it to determine whether laboratory test results are abnormal or not.

During the model training phase, we used \texttt{batch size} of 256, set the initial learning rate to $1\times10^{-4}$, and iterated the model for 20 epochs.

At the same time, since in clinical practice, a patient usually only measures a subset of the target laboratory indicators, there are a large number of missing labels in the data.These missing labels are represented by $-1$, and the one-hot label matrix we obtain is highly sparse.

To address this challenge, we used a customized masked binary cross-entropy loss function (Masked Binary Cross-Entropy Loss). In it, let $Y \in \{0, 1, -1\}^{N \times C}$ be the label matrix of a batch, $N=256$ is the batch size, $C=108$ is the total number of categories, and $-1$ represents that the inspection item is not detected. Let $\hat{L} \in \mathbb{R}^{N \times C}$ be the original logits output by the model. Our approach is to calculate the loss only for the observed labels (that is, labels with values of $0$ or $1$). We achieve this through a binary mask matrix $M \in \{0, 1\}^{N \times C}$, which is defined as follows:
$$
M_{i,j} =
\begin{cases}
1, & \text{if } Y_{i,j} \neq -1 \\
0, & \text{if } Y_{i,j} = -1
\end{cases}
$$
where $i$ and $j$ are the indices of the batch and category, respectively.

At the same time, we create a new label matrix $Y^{\prime} \in \{0, 1\}^{N \times C}$, replacing the missing values with $0$. Since the corresponding values of these positions in the mask $M$ are $0$, their final loss contribution will be nullified, so this replacement is safe.
$$
Y^{\prime}_{i,j} =
\begin{cases}
Y_{i,j}, & \text{if } Y_{i,j} \neq -1 \\
0, & \text{if } Y_{i,j} = -1
\end{cases}
$$

We use the standard binary cross entropy loss to calculate the element-wise loss.

The final total loss $\mathcal{L}_{total}$ is normalized by multiplying the element-wise loss with the mask $M$, summing the results, and dividing by the total number of valid labels in the batch:
$$
\mathcal{L}_{total} = \frac{\sum_{i=1}^{N} \sum_{j=1}^{C} M_{i,j} \cdot \mathcal{L}_{BCE}(\hat{L}_{i,j}, Y^{\prime}_{i,j})}{\sum_{i=1}^{N} \sum_{j=1}^{C} M_{i,j} + \epsilon}
$$
where $\epsilon$ is a very small value (here is set to $1 \times 10^{-8}$) to prevent division by zero in the rare case when all labels in the batch are invalid.

This approach ensures that the model's gradient updates are driven only by observed label-prediction pairs, enabling the model to robustly learn from incompletely annotated data.

\section*{ACKNOWLEDGMENTS}


Shenda Hong is supported by the National Natural Science Foundation of China (62102008, 62172018), CCF-Tencent Rhino-Bird Open Research Fund (CCF-Tencent RAGR20250108), CCF-Zhipu Large Model Innovation Fund (CCF-Zhipu202414), PKU-OPPO Fund (BO202301, BO202503), Research Project of Peking University in the State Key Laboratory of Vascular Homeostasis and Remodeling (2025-SKLVHR-YCTS-02).




\section*{DECLARATION OF INTERESTS}


The authors declare no competing interests.

\bibliography{reference}

\bigskip


\end{document}